\pdfoutput=1

\documentclass[11pt,dvipsnames,table,xcdraw]{article}

\usepackage[]{emnlp2021}

\usepackage{times}
\usepackage{latexsym}
\usepackage{multirow}               
\usepackage[T1]{fontenc}

\usepackage[utf8]{inputenc}

\usepackage{array}
\usepackage{tabularx}
\usepackage{microtype}
\usepackage{times}
\usepackage{epsfig}
\usepackage{graphicx}
\usepackage{amsmath}
\usepackage{amssymb}
\usepackage{bbm}
\usepackage{tablefootnote}
\usepackage{booktabs}
\usepackage{subcaption}
\usepackage{microtype}
\usepackage{paralist}
\usepackage{xcolor}
\usepackage{enumitem}
\usepackage{placeins}
\usepackage{geometry}  

\usepackage{comment}
\usepackage{xparse}
\NewDocumentCommand{\heng}
{ mO{} }{\textcolor{red}{\textsuperscript{\textit{Heng}}\textsf{\textbf{\small[#1]}}}}

\NewDocumentCommand{\lovish}
{ mO{} }{\textcolor{green}{\textsuperscript{\textit{lovish}}\textsf{\textbf{\small[#1]}}}}

\newcommand{\vmmee}{VM\textsuperscript{2}E\textsuperscript{2}}

\newcommand{\immee}{Image M\textsuperscript{2}E\textsuperscript{2}}

\title{Joint Multimedia Event Extraction from Video and Article}

\author{Brian Chen\Thanks{\hspace{.3em}Equal contribution.}\hspace{.3em}\textsuperscript{\textnormal{1}}, Xudong Lin\footnotemark[1]\hspace{.3em}\textsuperscript{\textnormal{1}}, Christopher Thomas\textsuperscript{\textnormal{1}}, Manling Li\textsuperscript{\textnormal{2}},\\
\textbf{Shoya Yoshida\textsuperscript{\textnormal{1}}, Lovish Chum\textsuperscript{\textnormal{1}}, Heng Ji\textsuperscript{\textnormal{2}}, Shih-Fu Chang\textsuperscript{\textnormal{1}}}. \\
  \textsuperscript{\textnormal{1}}Columbia University \textsuperscript{\textnormal{2}}University of Illinois at Urbana-Champaign \\

  \texttt{\{bc2754,xudong.lin,christopher.thomas,}\\
  \texttt{sy2905,lc3454,sc250\}@columbia.edu }\\
  \texttt{{\{manling2,hengji\}}@illinois.edu} \\
 }

\begin{document}
\maketitle

\begin{abstract}
Visual and textual modalities contribute complementary information about events described in multimedia documents. 
Videos contain rich dynamics and detailed unfoldings of events, while text describes more high-level and abstract concepts. However, existing event extraction methods either do not handle video or solely target video while ignoring other modalities. In contrast, we propose the first approach to jointly extract events from video and text articles. We introduce the new task of Video MultiMedia Event Extraction (V$\textrm{M}^2\textrm{E}^2$) and propose two novel components to build the first system towards this task. First, we propose the first self-supervised multimodal event coreference model that can determine coreference between video events and text events without any manually annotated pairs. Second, we introduce the first multimodal transformer which extracts structured event information jointly from both videos and text documents. 
We also construct and will publicly release a new benchmark of video-article pairs, consisting of 860 video-article pairs with extensive annotations for evaluating methods on this task. Our experimental results demonstrate the effectiveness of our proposed method on our new benchmark dataset. We achieve $6.0\%$ and $5.8\%$ absolute F-score gain on multimodal event coreference resolution and multimedia event extraction. 


\end{abstract}
\section{Introduction}

\begin{figure}[t]
  \includegraphics[width=1\columnwidth]{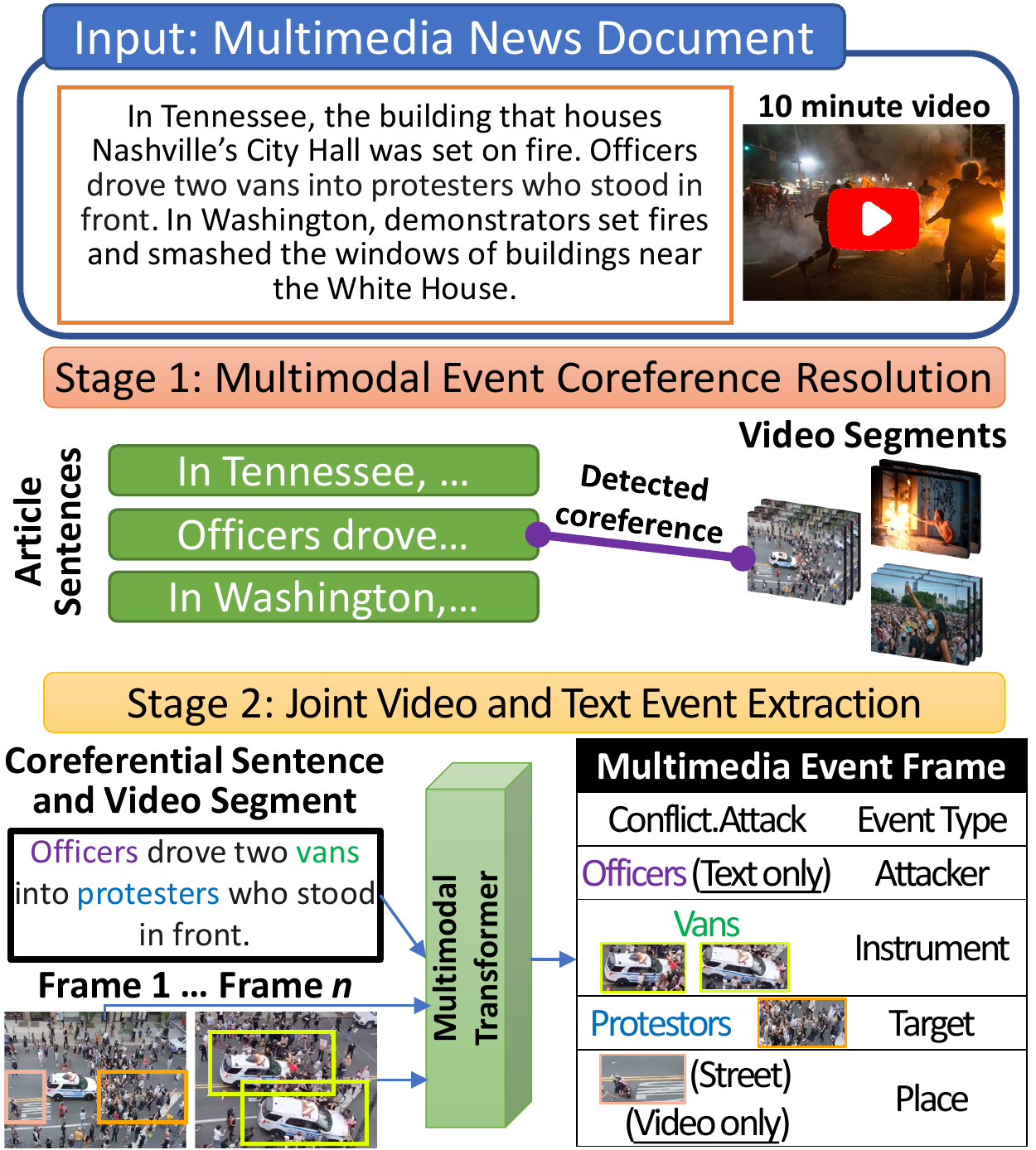}
\caption{We introduce the problem of video multimedia event extraction. Given a multimedia document containing a text article and a video, the goal is to jointly extract events and arguments. Our method first performs multimodal event coreference resolution to identify which sentences and video segments refer to the same event. Our novel multimodal transformer then extracts multimedia event frames from coreferential sentence and video segment pairs. Our method is able to resolve coreference and extract multimodal event frames more accurately than existing approaches.
}

\label{fig:concept}
\end{figure}

Traditional event extraction methods target a single modality, such as text, images, or videos. However, real-world multimedia (e.g.~online news) features content in multiple modalities which collectively convey a cross-modal narrative. As a consequence, components of events described by the document may lie jointly or solely in either the textual or visual modalities. By randomly watching 100 videos and associated articles from BBC Official YouTube Channel, we find that 45\% of videos contain event arguments that are not explicitly mentioned in the article. 


Event extraction is a well-studied problem in the natural language processing community \citep{nguyen-etal-2016-joint, sha2018jointly, liu-etal-2019-neural-cross, liu2020event}. Similarly, methods focusing on event argument extraction have likewise been proposed \citep{wang2019hmeae,wang2020neural}. However, all of these methods solely target the text modality and ignore the contribution of visual media. A related line of research has emerged in the computer vision community focusing on the extraction of purely visual events \citep{yatskar2016situation,li2017situation,mallya2017recurrent,pratt2020grounded}. While a few methods have sought to transfer visual knowledge from images to improve text-only event extraction \citep{zhang2017improving, tong2020image}, these do not detect multimodal events, whose arguments span multiple modalities.

\citet{li2020cross} propose a method for extracting multimodal events from text and images jointly. However, \citet{li2020cross}'s method does not handle videos. Extending \citet{li2020cross} to the video domain is non-trivial because localizing events in videos requires first identifying temporal boundaries of the event, which is a challenging vision problem in its own right \citep{pardo2021refineloc, huang2020improving, lin2019bmn}. Moreover, while \citet{li2020cross} transfer existing image and text event extraction resources to the multimodal domain, there are no datasets containing event argument localization in videos, thus \citet{li2020cross}'s method cannot be directly trained for multimodal text and video event extraction as it was for images.

We argue that multimodal event extraction from videos is important for several reasons. For one thing, images contain snapshots of events, but may not capture all arguments or participants of the event in a single snapshot. In contrast, videos often contain more action events and may reveal additional event arguments that can be extracted as events evolve over time that may be missing from any single frame.
Finally, we find some event argument \textit{roles} are hard to determine from single images, while video provides additional context which helps disambiguate the roles different arguments play in the event. 


In this paper, we propose the first model that extracts multimodal events and arguments from text and videos jointly. 
Specifically, we propose a new task called Video $\textrm{M}^2\textrm{E}^2$ (Video \textbf{M}ulti\textbf{M}edia \textbf{E}vent \textbf{E}xtraction). Given a document with an accompanying video, our goal is to jointly extract the events and argument roles appearing in both data modalities. 
Because of the lack of an existing dataset for this task, we introduce a new multimodal video-text 
dataset with extensive annotations covering event and argument role extraction, coreference resolution, and grounding of event arguments (bounding boxes).


We tackle this task in a two-stage manner: first we find a coreferential sentence-segment pair and then we jointly extract events from it. For multimodal event coreference resolution, we propose a self-supervised model to find video segment-sentence pairs describing the same event. These coreferential cross-modal pairs are then used to perform event classification and argument role labeling. To do so, we propose a novel multimodal transformer architecture which learns to perform event and argument role prediction jointly from video and text. We show that this system substantially outperforms unimodal approaches, while allowing us to discover event arguments lying solely in one modality. 




 
To summarize, we make the following contributions.
\begin{itemize}[nolistsep,noitemsep]
\item We propose the novel problem of video multimodal event extraction and contribute a high-quality benchmark dataset for this task containing extensive annotations of event types, event arguments and roles, argument grounding, and cross-modal coreference resolution of events in text and videos.
\item We propose a self-supervised training strategy which allows us to find coreferential sentence and video segment. 
\item We introduce a novel multimodal transformer architecture leveraging modality-specific decoders for joint text and video event and argument extraction.
\item We present extensive experimental results demonstrating that our proposed approach significantly outperforms both unimodal and multimodal baselines for event coreference resolution, event extraction, and argument role labeling.
\end{itemize}


\section{Related Work}

\noindent \textbf{Learning multimodal common space. }

Instead of learning representations in single modalities (text, visual), there have been various works that tried to learning representation from textual and visual modalities jointly and acquire a common space where the features from different modalities are directly comparable \citep{miech2019howto100m,miech2020end,chen2021multimodal}.
In the task such as weakly supervised grounding also tries to find a common space for text and visual where we can find the correct region given a text query \citep{akbari2019multi,zhang2020counterfactual,gupta2020contrastive}. These works usually learn in a weakly supervised manner where human-annotated image/video caption pairs were given.
In our multimodal event coreference resolution task, we try to learn in a self-supervised manner where only the video and its ASR were given.

\noindent \textbf{Text Event Extraction. }
Recognizing and extracting events in text is an important information extraction problem that has been thoroughly studied. Both document-level \citep{yang2018dcfee,Li2021doclevel} and sentence-level \citep{zeng2018scale} methods have been proposed. Classic work by \citet{ahn2006stages} and \citet{ji2008refining} leverage manually designed features for the task and formulate event extraction as a classification problem. More recent event extraction methods have leveraged neural models such as recurrent networks \citep{nguyen-etal-2016-joint, sha2018jointly}, convolutional networks \citep{chen-etal-2015-event}, graph networks \citep{liu-etal-2019-neural-cross,Zhang2021}, joint neural model~\cite{LinACL2020}, conditioned generation~\cite{Li2021doclevel} and transformers \citep{liu2020event} to automatically learn task-relevant features. 

A related line of work has focused on the problem of event argument extraction, where the goal is to predict event argument roles of entities in text to fill the roles of predicted event frames. \citet{wang2019hmeae} propose a hierarchical event argument extraction model leveraging modular networks to exploit argument role concept correlation. \citet{wang2020neural} propose a sampling-based method for jointly extracting events and arguments. Other methods have attempted to leverage zero-shot learning \citep{huang-etal-2018-zero} and weak supervision \citep{chen-etal-2017-automatically} to further improve performance on both event and event argument extraction.

While impressive progress has been made in recent years, all of these methods exclusively focus on text and forego the oftentimes complex and complementary information found in visual media. In contrast, we propose to extract both events and event arguments from both text and video. 

\noindent \textbf{Visual event extraction. }
Event recognition has also been studied by the computer vision community, where it is commonly termed ``situation recognition'' \citep{yatskar2016situation, pratt2020grounded}. Analogous to textual event extraction methods, the goal of visual situation recognition is detecting events occurring in an image, the objects and agents involved, and identifying their roles. Most work in this space \citep{yatskar2016situation,li2017situation,mallya2017recurrent,pratt2020grounded} relies on the FrameNet \citep{framenet} ontology derived from text which defines frames for each verb, along with semantic roles of arguments. 

Seminal work by \citet{yatskar2016situation} introduced the \textit{SituNet} dataset of images labeled with visual verbs and argument roles. Follow-up approaches have leveraged structured prediction mechanisms \citep{li2017situation,suhail2019mixture} and attention \citep{cooray2020attention} to further improve performance on \textit{SituNet}. \citet{pratt2020grounded} extend \textit{SituNet} with bounding box annotations of event arguments and introduce a model for localizing event arguments in images. None of these target the video domain as we do or perform multimodal event extraction.  More related to our work is \citet{Sadhu_2021_CVPR}, which introduces the video semantic role labeling dataset and task, where the target is to extract events and generate language description for arguments. Unlike \citet{Sadhu_2021_CVPR}, we propose to extract multimodal events and localize arguments, where components in the extracted event frame may appear in either modality.

\noindent \textbf{Multimodal Event Extraction.}
Some prior work has leveraged multimodal information for text-only event extraction. \citet{zhang2017improving} propose a method which learns to transfer visual knowledge from multimodal resources to text-only documents to improve event extraction. \citet{tong2020image} supplement existing event detection benchmarks with image data and show significant performance gains by leveraging multimodal information for trigger disambiguation.

Most relevant to our work is \citet{li2020cross}'s method which introduces the task of multimedia event extraction, where event frames are comprised of both visual and textual arguments. \citet{li2020cross} leverage single-modality training corpora and weak supervision to train a cross-modal method, without any annotations. Our work has several important differences from \citet{li2020cross}. First, we target the video modality, while \citet{li2020cross} target images. This problem is significantly more challenging because video event extraction requires understanding the rich dynamics in videos. 
Additionally, because no datasets of video event argument role localization exist, we can not directly borrow existing image event extraction resources like \citet{li2020cross}. 
Finally, we propose a novel multimodal transformer architecture for this task.



\section{\vmmee Dataset}
\subsection{Dataset Collection}
\label{subsec:data_collection}
We introduce the \vmmee dataset which labels (1) Multimodal event coreference (2) Events and argument roles from 860 video article pairs. 


\noindent \textbf{Event types.}
The Linguistic Data Consortium (LDC) has created document-level event ontology based on previous LDC-supported ontologies ERE and ACE. These have been made publicly availble online\footnote{\url{https://tac.nist.gov/tracks/SM-KBP/2018/ontologies/SeedlingOntology}}. The event types covered by the LDC ontology focus on issues related to disasters, attacks and activities from international news. We found that this ontology provides good coverage of many events found in world news and thus adopt it for our system.
Because not all event types in the ontology are visually detectable, we manually selected event types defined in the LDC ontology that are:
(1) Visually detectable: events that can be visually seen, and
(2) Frequent: events that have a frequency > 20 in our dataset. 
This resulted in a set of 16 event types, which we show in Table \ref{table:types}.
\begin{table}[t]
\centering
\resizebox{1\linewidth}{!}{
\begin{tabular}{|l|l|}
\hline
\multicolumn{2}{|c|}{\textbf{Event Type }}  \\
\hline
\hline
{CastVote (26)} & {Disaster.FireExplosion  {(60)}} \\
\hline
{Contact.Broadcast (359)} &  Life.Injure  {(78)} \\
\hline
{Contact.Correspondence (75)} & {Justice.ArrestJail (31)} \\ 
\hline
{Contact.Meet (196)} & {ManufactureAssemble (44)}\\
\hline
{Conflict.Attack (147)} & {Movement.Evacuation(23)}  \\
\hline
{Conflict.Demonstrate (242)} & {Movement.PreventPassage (43)} \\
\hline
{DamageDestroy (50)} &  {Movement.Transport (287)}   \\
\hline
{DetonateExplode (62)} &   {Transcation.ExchangeBuySell (36)} \\
\hline
\end{tabular}
}
\caption{Event types in \vmmee. 
Numbers in parentheses represent the counts of visual events. 
}
\label{table:types}
\end{table}
 The full event type and argument role definition are included in the supplementary.



\noindent \textbf{Candidate Video/Article Filtering.}
Given the 16 event types, we build a data collection pipeline. First, we use the event types and news source names as keywords to search on Youtube. We harvest from VOA, BBC, and Reuters. We choose these sources because we they are trustworthy and usually contain articles under the video such that the content is about the same event as the video. Second, we filter out videos that are longer than 16 minutes to avoid extra-long videos. Third, we check each video to make sure it contains at least one visual event. Starting from 1.2K videos, we end up with 860 video article pairs containing multimodal events. For the dataset, we will release the YouTube URLs that contains the video and article along with the annotations. We do not own the copyright of the video and the researcher shall use the data only for non-commercial research and educational purposes. More information about the Fair Use Notice will be included in the supplement.%


\subsection{Dataset Annotation Procedure}
In order to collect annotations, we perform the following steps for videos and text. First, annotators watch the entire video to identify all event instances in the video. Next, for each event instance, the temporal boundary, event type, and co-referential text event (if existent) is annotated and three keyframes within the temporal boundary are selected. Then, for each selected keyframe all arguments are identified. Finally, for each argument, the argument role type, entity type, and co-referential text event (if existent) is annotated.
We extensively annotate the videos and sampled keyframes with bounding boxes for argument roles to ensure none are missed.

\subsection{Quality control}

We train fourteen NLP and computer vision researchers to complete the annotation work with two independent passes. After annotation, two expert annotators perform adjudication. 
For the multimodal event coreference resolution, we sampled 10\% of annotations and reached an Inter-Annotator Agreement (IAA) of 84.6\%. For the event and argument role labeling, we sampled 10\% of annotations and reached an Inter-Annotator Agreement (IAA) of 81.2\%.


\subsection{Dataset statistics}
Overall, we annotated 852 multimodal event coreference links between video segments and sentences. \tablename~\ref{table:coref_stat} breaks down the annotations into relation categories: \textbf{1-to-1}, where one text event is only coreferential with a single video event, and \textbf{n-to-n}, where multiple text events and video events are corerefential.
We also provide data statistics for the event extraction and argument role annotations in \tablename~\ref{table:dataset_statistics}.

\begin{table}[!t]
\centering
\small
\setlength\tabcolsep{3pt}
\setlength\extrarowheight{2pt}
\begin{tabular}{ |c|c|c|c|c|c| } 
 \hline
 Type & \textbf{1-to-1} & \textbf{1-to-n} & \textbf{n-to-1} & \textbf{n-to-n} & \textbf{Total} \\ 
 \hline
 \hline
 Count & 202 & 104 & 260 & 286 & 852 \\ 
 \hline
\end{tabular}
\caption{Multimodal event coreference link types found in \vmmee.}
\label{table:coref_stat}
\end{table}

\begin{table}[!t]
\centering
\small
\begin{tabular}{|c|c|c|c|c|c|}
\hline
\multicolumn{2}{|c|}{\textbf{Document}} & \multicolumn{2}{c|}{\textbf{Event Mention}} & \multicolumn{2}{c|}{\textbf{Argument Role}} \\
\cline{1-6}
 Sentence  & Video & Textual & Visual & Textual & Visual  
\\ 
\hline
\hline
13,239 & 860 & 4,164 & 2,702 & 18,880 & 5,467 \\
\hline
\end{tabular}
\caption{Annotated \vmmee~data event and argument role statistics.} 
\label{table:dataset_statistics}
\vspace{-9pt}
\end{table}


\section{Method}

\begin{figure}[t]
 \resizebox{\columnwidth}{!}{
  \includegraphics[width=\columnwidth]{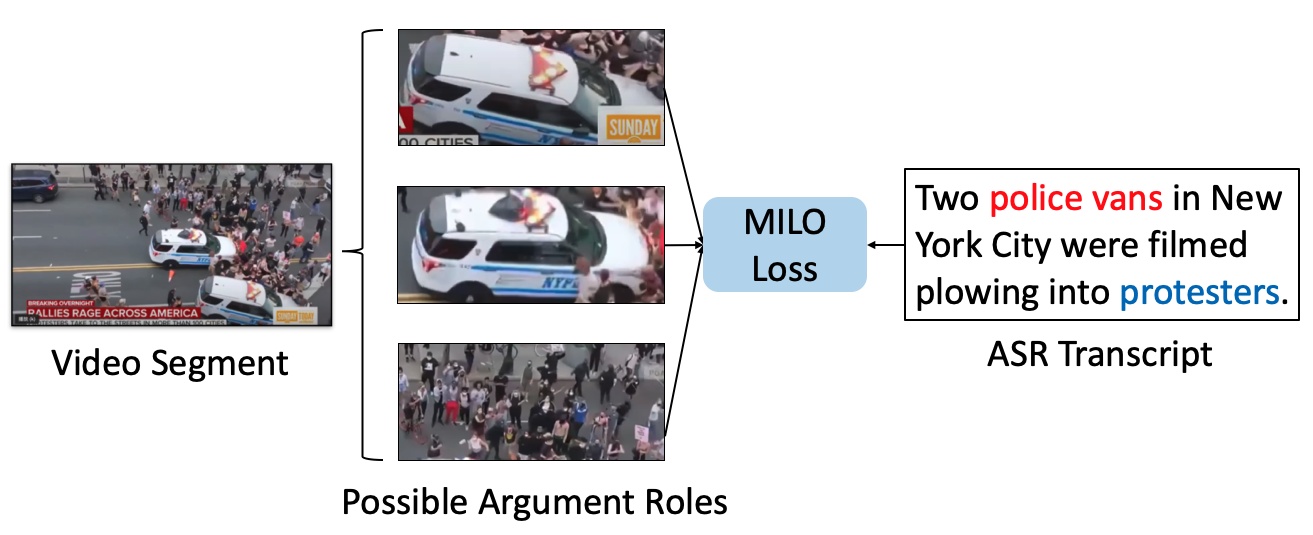}}
\caption{Self-supervised multimodal event coreference resolution by considering the possible argument roles that participate in the event.}
\label{fig:mm_coref}
\end{figure}

\subsection{Problem Formulation}
In Multimodal Event Coreference Resolution, given $M$ sentences and $N$ video segments in a multimedia document, the system is required to predict the coreference $c_{ij} \in \{0,1\}$ between a sentence $x_i$ and a video segment $y_j$. In Joint Multimodal Event Extraction and Argument Role Labeling, given a text sentence $x_i$ and a video segment $y_j$, the system is required to predict the multimodal event type $e$, the text mention $t_e$, the text mention $t_{a_k}$ and the bounding box $bbox_{a_k}$ for each argument role $a_k$.

\subsection{Multimodal Event Coreference Resolution}
 



We aim to learn a common space across the video and text modalities such that the embeddings across these modalities are close if they represent the same event.
This is a particularly challenging task since in an unannotated multimodal document, we don't 
know which video segment aligns with which article sentence.
Inspired by multimodal self supervised methods learning from instructional videos \cite{miech2019howto100m}, we learn the common space across the two modalities from our unannotated video clips using their auto-generated ASR transcripts as supervision. 
To accomplish this, we use a standard noise contrastive loss (NCE)~\cite{jozefowicz2016exploring} \emph{$\mathcal{L}_{NCE}$}:
\begin{equation}
\max_{f,g}\sum_{i=1}^n\log\left(\frac{e^{ f(x_i)^\top g(y_i)}}{e^{ f(x_i)^\top g(y_i)}+\sum\limits_{(x',y')\sim\mathcal{N}_i}\hspace*{-4mm}e^{ f(x')^\top g(y')}}\right) \nonumber
\end{equation}
where $x$ represents a sentence and $y$ a video clip.
$f$ and $g$ are the two learnable networks that project the two features into a common space. 
The loss learns to pull the positive pairs $(x_i,y_i)$ that co-occur in time while pushing mis-matched pairs in the batch away.

Additionally, we find the region information (arguments that participate in the event) to be crucial in finding coreferential events between video and text.
For example, when we see an \emph{Attack} event in the text, we might find the objects ``\texttt{van}'' or ``\texttt{protester}'' in the video to be important since they participate in the event as shown in Figure~\ref{fig:mm_coref}.
In order to learn such correspondences 
between text and object regions, we introduce the Multi-Instance Learning from Objects \emph{$\mathcal{L}_{MILO}$} loss:
\begin{equation}
\label{eq:objective}
\resizebox{0.95\linewidth}{!}{$
\max_{f,h}\sum_{i=1}^n\log\hspace*{-1mm}\left(\frac{\sum\limits_{(x,z)\in\mathcal{P}_i}e^{ f(x)^\top h(z)}}{\hspace*{-1mm}\sum\limits_{(x,z)\in\mathcal{P}_i}\hspace*{-2mm}e^{ f(x)^\top h(z)}+\hspace*{-5mm}\sum\limits_{(x',z')\sim\mathcal{N}_i}\hspace*{-4mm}e^{f(x')^\top h(z')}}\right)\hspace*{-1mm} \nonumber
$}
\end{equation}
where $z$ represents the regions in the video clip and $h$ is 
a projection layer.
Given a specific video instance $i$, $\mathcal{P}_i$ represents the \emph{positive} region/sentence candidate pairs (i.e.~the region and sentence co-occur in time, see Figure~\ref{fig:mm_coref}) while $\mathcal{N}_i$ represents the set of negative region/narration pairs that were sampled from different time frames.
The learning objective takes all possible region information into consideration by summing over all the pairs. The model learns in a multi-instance fashion to select the regions that are most important for multimodal event coreference resolution.
Our final multimodal coreference loss combines both global and local constraints:
\begin{equation}
\mathcal{L}_{mmcoref} = \mathcal{L}_{NCE} + \mathcal{L}_{MILO} \; \; \nonumber.
\end{equation}

\begin{figure*}[bpt]
\centering
 \resizebox{2\columnwidth}{!}{
  \includegraphics[width=\columnwidth]{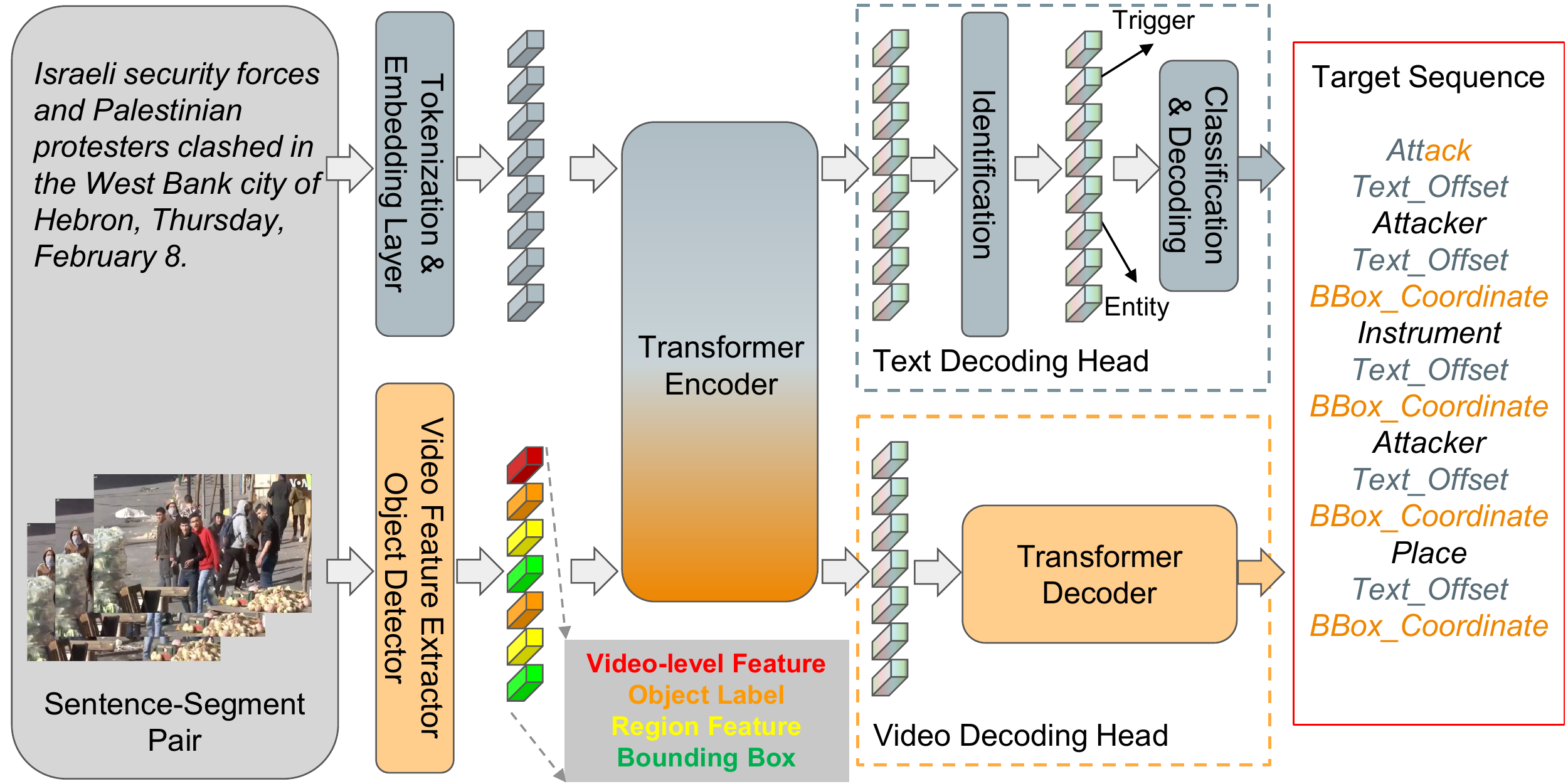}}
\caption{Multimodal transformer for joint event extraction and argument role labeling. In the target sequence, blue-gray and light orange are for textual and visual decoding heads, respectively.
}
\label{fig:mm_transform}
\end{figure*}

\subsection{Joint Multimodal Event Extraction and Argument Role Labeling}
Inspired by recent work \cite{lin2021vx2text} on leveraging multimodal transformers to jointly process text and visual information, we propose a joint multimodal transformer (JMMT) to extract events and arguments from a paired text sentence and video clip. The proposed JMMT has an encoder-decoder structure: the encoder extracts and fuses information from both modalities (text and video), while the decoder is more complex. The decoder consists of two heads: one for detecting trigger words, event types, and arguments from text, and the other for classifying video event types and predicting bounding boxes for visual arguments. With this joint encoder, JMMT can effectively leverage contextual information to extract events and label argument roles.

As shown in Figure~\ref{fig:mm_transform}, JMMT takes text and visual tokens as input. For text tokens, we follow \citet{raffel2019exploring} to embed text tokens. For visual tokens, we use four feature types to comprehensively represent both global and local information: 1) video-level features extracted from the whole video segment capture the global event context; 2) frame-level object labels produced by an object detector; 3) frame-level region features  extracted from bounding boxes detected by the object detector provide fine-grained argument information; 4) frame-level object coordinates also provided by the object detector for localization of arguments. Note that we sample $t$ frames and for each frame, we sample $k$ objects with the highest confidence scores. The text and visual tokens are then stacked as a sequence and input to the encoder for joint processing.

Our encoder and decoder are initialized from transformers pretrained on text corpora \cite{raffel2019exploring}. Our decoding head for text event extraction is borrowed from \citet{lin2020joint}'s state-of-the-art text event extraction model. For text decoding, we take encoder outputs as input and first merge these multimodal contextualized embeddings of word pieces to obtain a representation for each word in the input sequence. Then we process these word representations for identification, classification and decoding, following \citet{lin2020joint}. For the video decoding head, we leverage the decoder from \citet{raffel2019exploring}'s pretrained text transformer and cast the task as a sequence-to-sequence prediction problem. We set the target sequence as $\{e, a_1, bbox, a_2, bbox,...,a_n, bbox\}$, which begins with event type $e$ and then goes through each argument role $a_i$ to produce the bounding box coordinates $bbox$ on the sampled key frames. 

Each decoding head is supervised by its own loss term and the gradients are both back-propagated to the encoder. The text decoding head is supervised based on the objective $\mathcal{L}_{text}$ proposed in \citet{lin2020joint} and the video decoding head is trained using a standard teacher-forcing strategy with cross-entropy loss \citep{raffel2019exploring} $\mathcal{L}_{video}$. The overall objective is 
\begin{equation}
    \mathcal{L}_{JMMT} = \mathcal{L}_{text} + \mathcal{L}_{video}. \nonumber
\end{equation}
In this way, the proposed JMMT can effectively fuse multimodal information and jointly extract events and arguments.

\section{Experiments}

\subsection{Dataset}
\noindent \textbf{Event coreference resolution.}
Our model is trained on our unannotated dataset, which contains 3K videos and corresponding automatically generated speech transcriptions.
We test our model on the annotated dataset, which contains 860 videos and their articles from YouTube. 

\noindent \textbf{Event extraction and argument role labeling.}
We split the annotated 860 video-article pairs into 645 and 215 for training and testing, respectively. To focus on joint multimedia event extraction, we sample all the coreference segment-sentence pairs for training and evaluation.

\subsection{Evaluation Setting}
\noindent \textbf{Event coreference resolution.}
We evaluate our model on the annotated event coreference data by predicting whether every possible sentence-video segment pair from the same multimodal document is coreferential or not. 
We perform feature similarity between the text and video features within the learned joint space and predict the pair as coreferential if their similarity surpasses a threshold.
We adopt traditional link prediction metrics, i.e.~precision, recall, \textit{F}\textsubscript{1}, and accuracy for evaluation.



\newcommand{\tabincell}[2]{\begin{tabular}{@{}#1@{}}#2\end{tabular}}

\begin{table*}[t]
\footnotesize
\centering
\setlength\tabcolsep{1pt}
\setlength\extrarowheight{1pt}
\begin{tabular}{|c|c|c c c|c c c|c c c|c c c|c c c|c c c|}
\hline
\multirow{3}{*}{{\textbf{Input}}} & \multirow{3}{*}{\textbf{Model}}
& 
\multicolumn{6}{c|}{\textbf{Text Evaluation}} & \multicolumn{6}{c|}{\textbf{Video Evaluation}} & \multicolumn{6}{c|}{\textbf{Multimedia Evaluation}}
\\ \cline{3-20}
& & \multicolumn{3}{c|}{\textbf{Event Mention}} & \multicolumn{3}{c|}{\textbf{Argument Role}} & \multicolumn{3}{c|}{\textbf{Event Mention}} & \multicolumn{3}{c|}{\textbf{Argument Role}} & \multicolumn{3}{c|}{\textbf{Event Mention}} & \multicolumn{3}{c|}{\textbf{Argument Role}}  
\\ \cline{3-20}
& & \textbf{$P$ }  & \textbf{$R$} & \textbf{$F_{1}$} & 
\textbf{$P$} & \textbf{$R$} & \textbf{$F_{1}$}  & 
\textbf{$P$} & \textbf{$R$} & \textbf{$F_{1}$}  & 
\textbf{$P$} & \textbf{$R$} & \textbf{$F_{1}$}  & 
\textbf{$P$} & \textbf{$R$} & \textbf{$F_{1}$}  & 
\textbf{$P$} & \textbf{$R$} & \textbf{$F_{1}$}  
\\ \hline
\multirow{1}{*}{~{\textbf{Text}}} 
& OneIE 
& 38.5 & 52.1 & 44.3 & 16.6 & 21.8 & 18.8 
& -  & - & - & - & - & -  
& 38.5 & 52.1 & 44.3 & 16.6 & 21.8 & 18.8   \\ 
\hline
\multirow{2}{*}{{\textbf{Video}}} 
& JSL
&  - & - & - & - & - & - 
& 24.1 & 17.1 & 20.0 & 2.2 & 2.8 & 2.4  
& 24.1 & 17.1 & 20.0 & 2.2 & 2.8 & 2.4  \\
& JMMT\textsubscript{Video}
&  - & - & - & - & - & - 
& 26.6 & 29.2 & 27.8 & 8.9 & 10.1 & 9.5
& 26.6 & 29.2 & 27.8 & 8.9 & 10.1 & 9.5 \\
\hline
\multirow{2}{*}{{\textbf{Multimedia}}} 
& WASE
& 33.6 & 53.8 & 41.4 & 15.2 & 22.1 & 18.0
& 20.4 & 14.0 & 16.6 & 2.8 & 1.3 & 1.7 
& 34.0 & 54.0 & 41.8 & 15.3 & 22.1 & 18.1  \\ 
& JMMT
& 39.7 & 56.3 & \textbf{46.6} & 17.9 & 24.3 & \textbf{20.6} 
& 32.4 & 37.5 & \textbf{34.8} & 9.2 & 10.6 & \textbf{9.9} 
& 41.2 & 56.3 & \textbf{47.6} & 18.8 & 24.7 & \textbf{21.3} \\ 
\hline
\end{tabular}
\vspace{-0.5em}
\caption{Event and argument extraction results (\%). We evaluate three categories of models in three evaluation settings. By jointly leveraging multimodal context, JMMT significantly improves multimedia event extraction from video segments and sentences. }
\label{table:result_m2e2}
\vspace{-0.5em}
\end{table*}

\noindent \textbf{Event extraction and argument role labeling.}
We evaluate models on text-only, video-only, and multimedia event mentions in the \vmmee dataset.
We follow the common event extraction metrics, i.e.~precision, recall, and \textit{F}\textsubscript{1}.
For a text event mention, we follow~\citet{li2020cross} to only consider it as correct if its trigger offsets and event type both match a reference trigger. Similarly, a textual argument is only considered as correct when its offsets, event type, and role type all match a reference argument. Analogously, a video event mention is considered correct if its segment and event type match a reference segment. A video argument is considered correct if its localization, event type and role type matches a reference argument. A visual argument is correctly localized if its Intersection over Union (IoU) with the ground truth bounding box is greater than 0.3. Finally, a multimedia event mention is considered correct if its event type and trigger offsets (or the video segment) match a reference trigger (or the reference segment). Arguments of multimedia events with either a correct textual or visual argument mention are considered correct. 

\subsection{Baseline methods}
\noindent \textbf{Event coreference resolution.}
We compare our method against several self-supervised models that learn a joint visual text space. Specifically, HowTo100m \citep{miech2019howto100m} learn a joint video-text space using a max-margin ranking loss \citep{karpathy2014deep}. NCE loss \citep{jozefowicz2016exploring} trains a classifier to discriminate between real instances and a generated noise distribution.  MIL-NCE \cite{miech2020end} further extends NCE by explicitly considering the misalignment of the video segment and ASR transcript to design a multi-instance loss. We do not compare to retrieval methods that require fine-tuning. \\
\noindent \textbf{Event extraction and argument role labelling.}
1) \noindent \textbf{Text-only baseline:} We re-implement a state-of-the-art method, OneIE~\citep{lin2020joint}. For a fair comparison, we use the same text encoder \citep{raffel2019exploring} as our JMMT. 2) \noindent \textbf{Video-only baseline:} As no existing method addresses the problem of event extraction and argument role labeling from videos, we adopt the state-of-the-art method for grounded image event extraction, JSL~\citep{pratt2020grounded}, to extract events and arguments from each annotated key frame. 3) \noindent \textbf{Multimedia baseline:} As previous multimedia event extraction methods only consider image-text pairs, we borrow one of the best performing models on $\textrm{M}^2\textrm{E}^2$ \cite{li2020cross}, WASE \cite{li2020cross}, as our baseline for multimedia event extraction. Note that we rebuild WASE to extend from its ontology to our ontology.


\subsection{Implementation details} 
For the visual branch of the multimodal event corerefence resolution model we follow \citet{miech2019howto100m} and use pre-trained 2D features from a ResNet-152 model~\cite{he2016deep} trained on ImageNet \cite{deng2009imagenet} and 3D features from a ResNeXt-101 model~\cite{hara2018can} trained on Kinetics~\cite{carreira2017quo}.
For the textual branch, a GoogleNews pre-trained Word2vec model~\cite{mikolov2013efficient} provides word embeddings, followed by a max-pooling over words in a given sentence to extract a sentence embedding. We use Faster R-CNN \cite{ren2015faster} pre-trained on the Visual Genome dataset \cite{krishna2017visual} as our object detector.
For selecting the number of objects, we sort by the confidence score of each object and select the top 5 as possible argument roles. Also, we uniformly sample 3 frames in each video segment and end up with 15 objects for each segment. More details about the object selection can be found in the supplementary material.

For event extraction and argument role labeling, we use the same video-level feature and object detector. We use T5-base \cite{raffel2019exploring} with pre-trained weights provided in HuggingFace~\cite{wolf2019huggingface} for initialization. More details are in the supplementary material. 

\begin{figure*}[t]
\centering
 \resizebox{2\columnwidth}{!}{
  \includegraphics[width=\columnwidth]{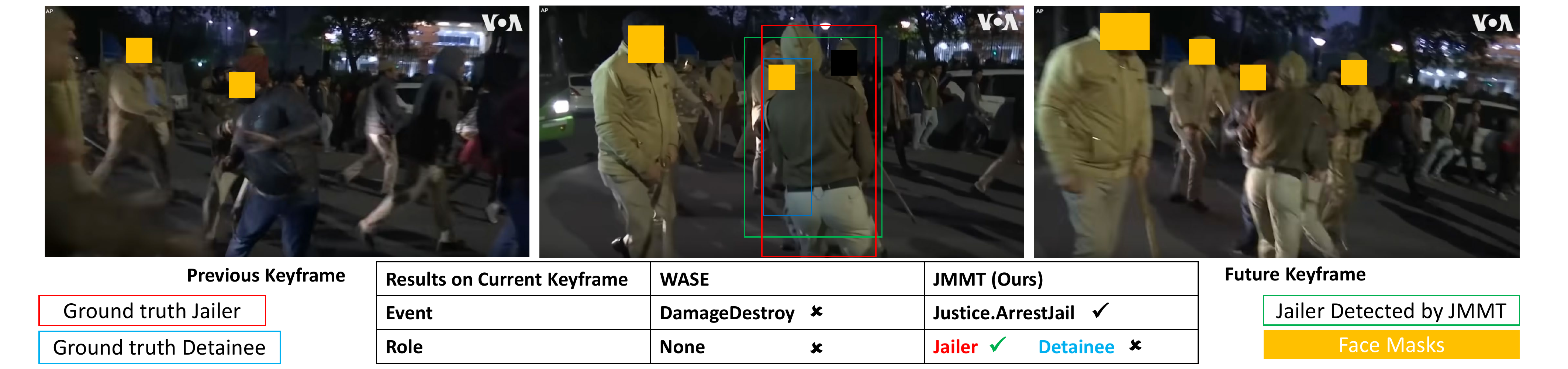}}
\caption{Visualization of event extraction results on one video segment. We mask faces (orange boxes) for privacy.} 
\label{fig:m2e2_visual}
\end{figure*}

\begin{table}[t]
        \centering
        \resizebox{\columnwidth}{!}{
		 \begin{tabular}[t]{@{}lcccccc@{}}
			\toprule
Method & Visual Model & TR & \textbf{$P$ }  & \textbf{$R$} & \textbf{$F_{1}$} & \textbf{$Acc$} \\
\hline
HowTo100M   & R152+RX101 & N   &  32.2  & 62.8 & 44.3 &  55.2  \\
NCE      & R152+RX101 & N &  35.5  & 68.3 & 45.5 & 47.5    \\
\textbf{Ours} & R152+RX101 & N &  \textbf{38.4}  &\textbf{ 76.4}  & \textbf{51.5} & \textbf{59.6}   \\
\midrule
MIL-NCE  & S3D-G & Y & 37.8 & 75.0 & 50.6 &   59.2 \\
\midrule
			\bottomrule
		\end{tabular}
		} 
		
		\caption{Multimodal event coreference resolution results. Our method outperforms all baselines, including one with a more powerful and trainable visual backbone (indicated by TR).}
		\label{tab:mm_coref}
		\vspace{-0.35cm}
\end{table}

\begin{figure}[t]
 \resizebox{\columnwidth}{!}{
  \includegraphics[width=\columnwidth]{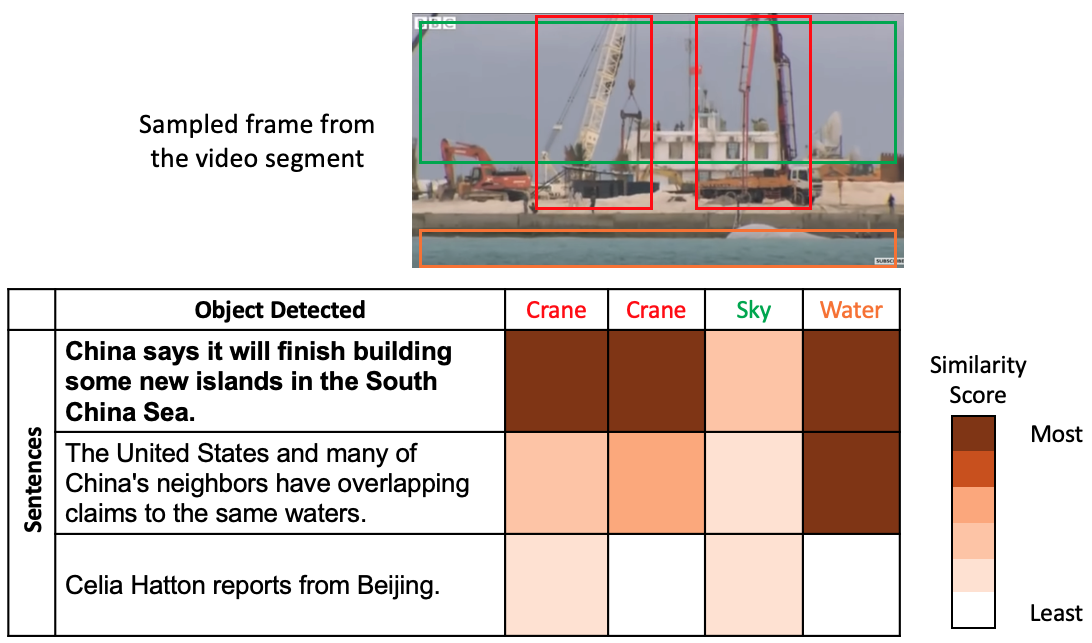}}
\caption{Event coreference resolution visualization. 
The bold sentence is correctly selected as coreferential within the article by the model.}
\label{fig:mmcoref_visual}
\end{figure}

\subsection{Quantitative Performance}
\noindent \textbf{Event coreference resolution.}
We first examine the results of the multimodal event coreference resolution task in Table~\ref{tab:mm_coref}.
All the methods we compare share the same text feature extractor. For visual feature extraction, HowTo100M, NCE, and our method apply ResNet-152 (R152) and ResNeXt-152 (RX101) followed by \citet{miech2019howto100m}. MIL-NCE uses a more advanced video feature extraction backbone, S3D-G \citep{xie2017rethinking}.
Our model significantly outperforms all previous methods using the same architecture, as well as those models with a trainable (TR) and more powerful visual backbone  \citep{miech2020end}.

\noindent \textbf{Event extraction and argument role labeling.}
The proposed JMMT significantly improves the event extraction performance over baseline methods as shown in Table~\ref{table:result_m2e2}. Compared to text-only OneIE or video-only JSL baselines, the JMMT produces at most \textbf{74\%} relative gain in event extraction, which demonstrates the importance of leveraging multimodal information for understanding complex events. Compared to previous methods on image-text multimedia event extraction, the superior performance of JMTT verifies 1) the effectiveness of the powerful transformer model for multimodal information fusion and 2) the importance of modeling dynamics in videos.




\subsection{Qualitative Analysis}
We visualize results from our event coreference resolution model in Fig \ref{fig:mmcoref_visual}. We observe that the model correctly selects the most appropriate sentence for a given video segment. Also, we find that the model learns to associate object regions to the words in the sentence. For example, the first sentence had a high similarity score with the object \emph{'Crane'} since it mentioned \emph{'building some new island'}. 

We also visualize results of event extraction. As shown in Fig.~\ref{fig:m2e2_visual}, the center frame is very hard to extract events from due to the occlusion of arguments. With only this image as input, WASE fails to extract events and arguments. However, our JMMT successfully recognizes the event and detects ``Jailer'' in the image with the help of video-level dynamics and the context of the previous and next frames. This example illustrates both the importance and difficulty of multimedia event extraction from videos and articles. We also observe that our JMMT fails to recognize the ``Detainee'' because of occlusion. This indicates the possibility of leverage entity tracking to further improve V$\textrm{M}^2\textrm{E}^2$ which we leave as future work.

\section{Limitation}

\noindent\textbf{Dataset.} Our method was based on our collected dataset, which might contain unintended societal, gender, racial, and other biases when deploying models trained on this data. Also, our problem formulation assumes the video and article are about the same topic. This assumption leads our method to work on news videos and instructional videos. If we didn't constrain the videos to these genres, we might collect videos without articles or videos with unrelated articles such as music videos and animated videos on YouTube.

\noindent\textbf{Evaluation.} 
Our proposed pipeline could be combined as a two-step approach, starting from raw videos and articles and then acquiring both modalities' event and argument roles. However, in our evaluation, we only evaluate the argument roles on annotated keyframes since we don't have the annotation for every video frame due to the expense of annotation. End-to-end evaluation for all frames is not practical because the chosen frames from the multimodal event coreference resolution model are not guaranteed to be the ground truth frames on which we have annotations. Consequently, we cannot evaluate the predictions of our multimodal event extraction model (stage 2) when we use predicted frames as input since we do not have annotations on the frames on which predictions are made (thus, the results in those frames could be correct or incorrect).
\section{Conclusions}
We have introduced a novel task \vmmee - given a video with a paired article, our first goal is to find coreferenced events across modalities. Also, our task requires extracting the event type and argument roles from both modalities. 860 video-article pairs were labeled to support this task.
We developed a novel self-supervised multimodal network that learns a common embedding space by processing local (object region) and global (video level) semantic relationships to perform multimodal event coreference resolution.
In addition, we present a new architecture JMMT that jointly extracts events and arguments from both modalities using an encoder-decoder-based multimodal transformer.
Our extensive experiments on multiple settings show that considering region information and a joint transformer for both modalities is essential for good performance on the two subtasks in \vmmee.
Our dataset collection pipeline and approach can be extended to more scenarios such as instructional videos and other videos that contain video-article pairs for extracting multimodal events across both modalities.

\section*{Acknowledgement}
This research is based upon work supported by U.S. DARPA KAIROS Program No. FA8750-19-2-1004, and U.S. DARPA AIDA Program No. FA8750-18-2-0014. The views and conclusions contained herein are those of the authors and should not be interpreted as necessarily representing the official policies, either expressed or implied, of DARPA, or the U.S. Government. The U.S. Government is authorized to reproduce and distribute reprints for governmental purposes notwithstanding any copyright annotation therein.

\bibliography{anthology}
\bibliographystyle{acl_natbib}

\clearpage
\title{Supplementary for \\Joint Multimedia Event Extraction from Video and Article}

\maketitle
\appendix

\noindent The appendix is organized as follows: \\
\textbf{A.} We include more experimental settings.  \\
\textbf{B.} We demonstrate our proposal generation component for selecting video segments with events.  \\
\textbf{C.} We list argument roles for each event type.\\
\textbf{D.} We provide more details of our baseline. \\
\textbf{E.} We show our annotation interface for the dataset.
\textbf{F.} We state our Fair Use Notice.

\section{Implementation details} 
\noindent\textbf{Multimodal event coreference resolution. }
For each feature extraction branch (text, video, object), we apply separate fully-connected layer and a gated unit for projection to common space. 
We use an Adam optimizer~\cite{kingma2015adam} with a learning rate of $1\mathrm{e}{-4}$. The batch size is set to 256 video clips.
The model is trained for 50 epochs on one NVIDIA TITAN RTX for about 2 hours. We further split the 860 video data into 200 video article pairs for the validation set and 660 for testing the performance. The parameter search for the threshold was done in the validation set by selecting the highest F1 score, and the similarity score above 0.13 will be viewed as positive pairs for prediction.

\noindent\textbf{Multimodal event extraction and argument role labeling.}
For video-level features and region features, we separately use a fully-connected layer to project them into $768$-D space to be aligned with text embeddings. We directly use text embedding layer to embed bounding box coordinates. For the text decoding head, we borrow implementations from the official implementation\footnote{http://blender.cs.illinois.edu/software/oneie} of OneIE and use the same hyper-parameters. The video decoding head uses Beam Search for decoding in inference, with a beam width of 5. 
During training and evaluation, we sample annotated $t=3$ frames and extract $k=15$ objects for each frame. We use a batch size of 6 examples per GPU, and distribute the training over 4 NVIDIA V100 GPUs. We use Adam with a learning rate of $1\mathrm{e}{-4}$ to optimize our models. We train our models for 150 epochs.

\section{Event Proposal Generation}
To acquire the temporal boundary of the video event, we use the Boundary Sensitive Network \cite{lin2018bsn} for temporal proposal generation in the video clips. We fine-tune the network with the \vmmee training set to better capture the action semantics within the dataset. Table \ref{tab:proposals} shows the proposal generation results for \vmmee test set. Similar to \cite{lin2018bsn, su2020bsn++}, we evaluate the improvement in the ability of BSN to generate proposals which have high temporal overlap with ground truth proposals. To quantify this improvement, we measure the recall (AR) over multiple temporal-IoU thresholds (0.5 to 0.95 with an increment of 0.05) for a fixed number of proposals(N). We also measure the area under(AUC) average recall(AR) at different number of proposals(N) curve.  

Although we use ground truth proposals for event extraction and argument role labelling section of the experiments in the current work, our method can be extended to work with automatically generated proposals. Hence, our method combined with any proposal generation technique, can be considered as an end-to-end solution to multimedia event extraction given a video-article pair.


\begin{table}[!h]
    \centering
    \begin{tabular}{|c|c|c|c|}
    \hline
    \multirow{2}{*}{\textbf{Training}}  & \multicolumn{2}{c|}{\textbf{AR}} & \multirow{2}{*}{\textbf{AUC}}\\ \cline{2-3} 
    & @1 & @100 & \\ 
    \hline
    ActivityNet  & 0.11  & 0.52 & 38.52\\
    ActivityNet + \vmmee  &  0.18  & 0.67 & 54.94 \\
    \hline
    \end{tabular}
    \caption{Fine-tuning the BSN pipeline with \vmmee shows significant improvement in proposal generation and retrival performance. }
    \label{tab:proposals}
\end{table}

\begin{table*}[tbhp]
\small
\centering
\resizebox{1\linewidth}{!}{
\begin{tabular}{|l|l|}
\hline
\textbf{Event Type} & \textbf{Argument Role}  \\
\hline
CastVote & Voter,Candidate,Ballot,Result,Place  \\  
\hline
Contact.Broadcast  & Communicator, Recipient, Instrument, Topic, Place  \\
\hline
Contact.Correspondence  & Participant, Instrument, Topic, Place \\
\hline
Contact.Meet  & Participant, Topic, Place  \\
\hline
Conflict.Attack  & Attacker, Target, Instrument, Place \\
\hline
Conflict.Demonstrate  & Demonstrator, Demonstrator, VisualDisplay, Topic, Target, Place \\
\hline
DamageDestroy & Damager, Artifact, Instrument, Place \\
\hline
DetonateExplode & Attacker, Target, Instrument, ExplosiveDevice, Place \\
\hline
Disaster.FireExplosion & FireExplosionObject, Instrument, Place \\
\hline
Life.Injure & Victim, Injurer, Instrument, BodyPart, MedicalCondition, Place \\
\hline
Justice.ArrestJail & Jailer, Detainee, Crime, Place  \\
\hline
ManufactureAssemble & ManufacturerAssembler, Artifact, Components, Instrument, Place   \\
\hline
Movement.Evacuation & Transporter, PassengerArtifact, Vehicle, Origin, Destination  \\
\hline
Movement.PreventPassage & Transporter, PassengerArtifact, Vehicle, Preventer, Origin, Destination \\
\hline
Movement.Transport & Transporter, PassengerArtifact, Vehicle, Origin, Destination  \\
\hline
Transcation.ExchangeBuySell & Giver, Recipient, AcquiredEntity, PaymentBarter, Beneficiary, Place   \\
\hline
\end{tabular}
}
\caption{Event types and argument roles in \vmmee.
}
\label{table:types_all}
\end{table*}

\section{Event type}
The event type along with its argument roles are shown in Table \ref{table:types_all}. We followed The Linguistic Data Consortium (LDC) ontology defined for the AIDA program. These have been made publicly availble online\footnote{\url{https://tac.nist.gov/tracks/SM-KBP/2018/ontologies/SeedlingOntology}}.


\section{Multimodal event extraction and argument role labeling Baselines}

SWiG (Situations with Grounding) dataset provides the annotations corresponding to the visually groundable verbs and the nouns associated with them. To evaluate the JSL\cite{lin2020joint} model on \vmmee dataset, we map the SWiG verb classes onto the \vmmee event classes as described in Table \ref{tab:swig_mapping}. Note that some classes in SWiG do not have any verb corresponding to the \vmmee event. Hence, these events are never predicted by the JSL model. For fair comparison, we calculate the precision and recall with respect to the remaining classes only.

In a similar manner, we reformulate the mappings used in WASE \cite{li2020cross} to extend the ontology of \immee \cite{li2020cross} to the \vmmee ontology and retrain WASE as our baseline.

\begin{table*}[]
    \small
    \centering
    \resizebox{1\linewidth}{!}{
    \centering
    \begin{tabular}{|l|p{25em}|}
    \hline
    \textbf{\vmmee Event Type}  & \textbf{SWiG Verb Class}\\ \hline
    CastVote  & Voting  \\ \hline
    Contact.Broadcast & Speaking   \\ \hline
    Contact.Correspondence & Calling, Dialing, Phoning, Telephoning \\ \hline
    Contact.Meet  & Communicating, Interviewing, Talking, Discussing, Shaking  \\ \hline
    Conflict.Attack & Attacking, Punching, Kicking, Striking, Shooting   \\ \hline
    Conflict.Demonstrate & Protesting, Marching, Displaying, Gathering  \\ \hline
    DamageDestroy & Breaking, Destroying   \\ \hline
    DetonateExplode & -\\ \hline
    Disaster.FireExplosion & Flaming, Erupting, Burning    \\ \hline
    Life.Injure & - \\ \hline
    Justice.ArrestJail & Detaining, Restraining, Arresting  \\ \hline
    ManufactureAssemble & Assembling \\ \hline
    Movement.Evacuation & -   \\ \hline
    Movement.PreventPassage & Blocking, Guarding\\ \hline
    Movement.Transport & Driving, Boating, Disembarking, Landing, Piloting, Steering, Taxiing, Commuting, Riding, Boarding, Biking  \\ \hline
    Transaction.ExchangeBuySell &  Paying, Selling \\ \hline
    \end{tabular}
    }
    \caption{Mapping used to convert the SWiG verbs to \vmmee events. Note that 3 events do not have any mapping. We do not evaluate the JSL baseline over these events. }
    \label{tab:swig_mapping}
\end{table*}

\section{Annotation interface}

\begin{figure*}[bpt]
 \resizebox{2\columnwidth}{!}{
  \includegraphics[width=\columnwidth]{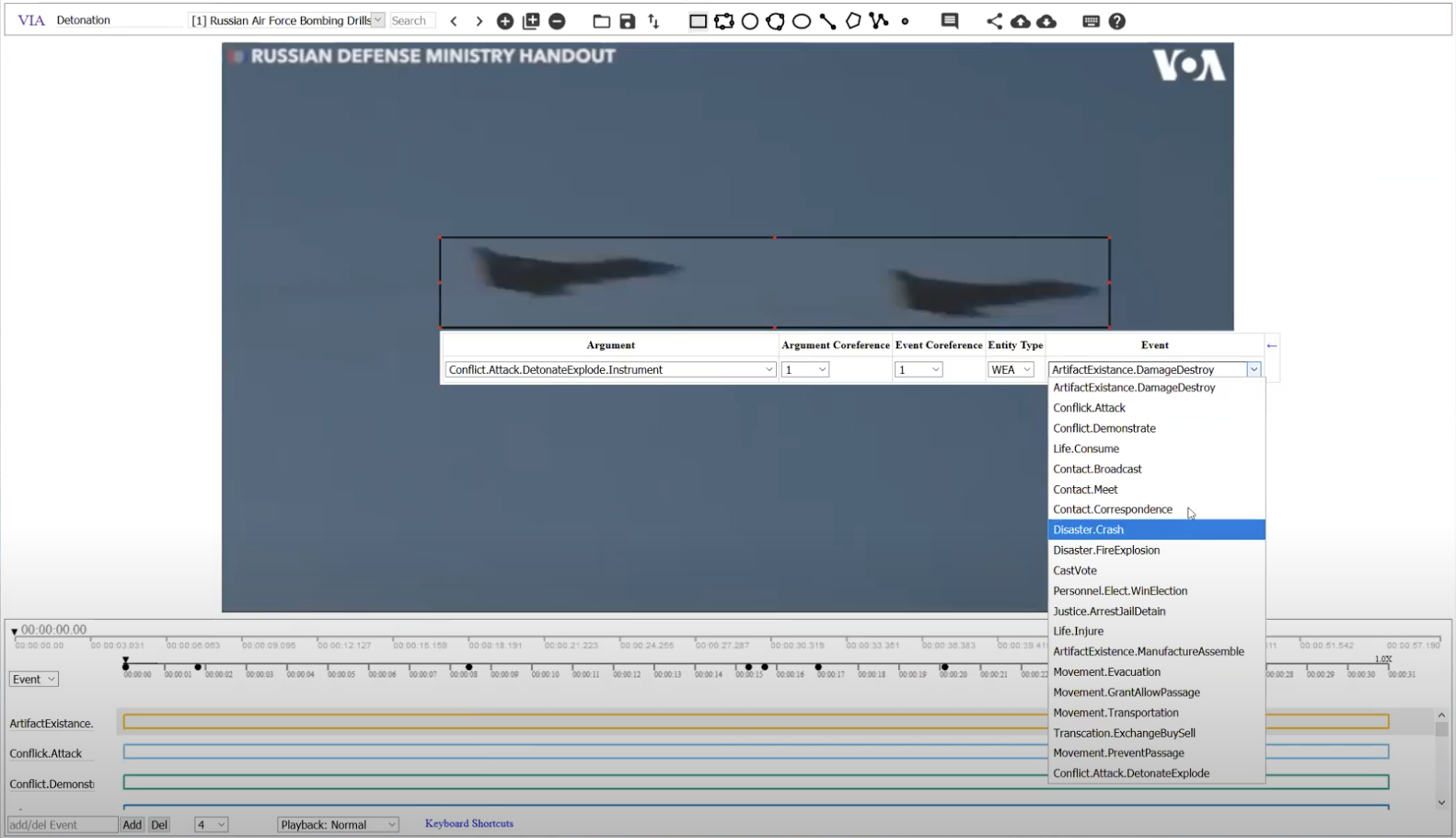}}
\caption{Annotation interface of the video. We annotate the event temporal of each video event. Also, we will annotate the multimodal event coreference between the video event and text event.
For the argument role, we select 3 frames to annotate the bounding box. }
\label{fig:anno}
\end{figure*}

Our data annotation interface for video is shown in Figure \ref{fig:anno}. Each annotator needs to walk through the whole video and corresponding articles. As shown in the figure, we have a list of event types for the annotators to label the start time and end time. The same event can appear multiple times in the same video. We also allow overlap between different events.

\section{Fair Use Notice}
Our dataset and this paper contain copyrighted material the use of which has not always been specifically authorized by the copyright owner. We make such material available in an effort to advance understanding of technological, scientific, and cultural issues. We believe this constitutes a `fair use' of any such copyrighted material as provided for in section 107 of the US Copyright Law. In accordance with Title 17 U.S.C. Section 107, the materials in this paper are distributed without profit to those who have expressed a prior interest in receiving the included information for non-commercial research and educational purposes. If you wish to use copyrighted material from this paper or in our dataset for purposes of your own that go beyond non-commercial research and academic purposes, you must obtain permission directly from the copyright owner.



\end{document}